\newcommand{\CLASSINPUTtoptextmargin}{37mm}
\newcommand{\CLASSINPUTbottomtextmargin}{19.1mm}
\newcommand{\CLASSINPUTinnersidemargin}{19.1mm}
\newcommand{\CLASSINPUToutersidemargin}{13.1mm}
\documentclass[a4paper,conference]{IEEEtran}
\usepackage{cite}
\usepackage{graphicx}
\usepackage{subfigure}
\usepackage{forest}
\usetikzlibrary{angles,shadows,arrows}
\usepackage{changes} 
\usetikzlibrary{arrows,arrows.meta,decorations.pathmorphing}
\begin{document}

\bstctlcite{IEEEexample:BSTcontrol}

\title{\vspace{6mm}Design Space of Behaviour Planning for Autonomous Driving}

\author{
\IEEEauthorblockN{Marko Ilievski, Sean Sedwards, Ashish Gaurav, Aravind Balakrishnan, Atrisha Sarkar\\Jaeyoung Lee, Fr{\'e}d{\'e}ric Bouchard, Ryan De Iaco, and Krzysztof Czarnecki}\\
\IEEEauthorblockA{University of Waterloo, Canada}
}

\maketitle

\begin{abstract}
We explore the complex design space of behaviour planning for autonomous driving. Design choices that successfully address one aspect of behaviour planning can critically constrain others. To aid the design process, in this work we decompose the design space with respect to important choices arising from the current state of the art approaches, and describe the resulting trade-offs. In doing this, we also identify interesting directions of future work. 
\end{abstract}

\begin{IEEEkeywords}
Autonomous Driving, Behaviour Planning, Motion Planning, Design Space.
\end{IEEEkeywords}

\section{Introduction}\label{sec:introduction}
In this work we consider the design space~\cite{MYBM91} of behaviour planning---high level decision making---for autonomous driving.
To simplify the design process, we decompose the design space into three principal axes of design choices, based on our practical experience~\cite{autonomoose100k} and with reference to the current state of the art.
Within each axis, we discuss the inevitable qualitative trade-offs that exist and review the relevant literature.
We illustrate our decomposition using feature diagrams~\cite{KCHN90}. 
%
In doing this, we identify potentially interesting areas of research within the behaviour planning design space.
The motivation of our decomposition is as follows.

\begin{figure}[t]
\centering
\subfigure[Raw data: Lidar point cloud.]{
\frame{\includegraphics[width=.225\textwidth]{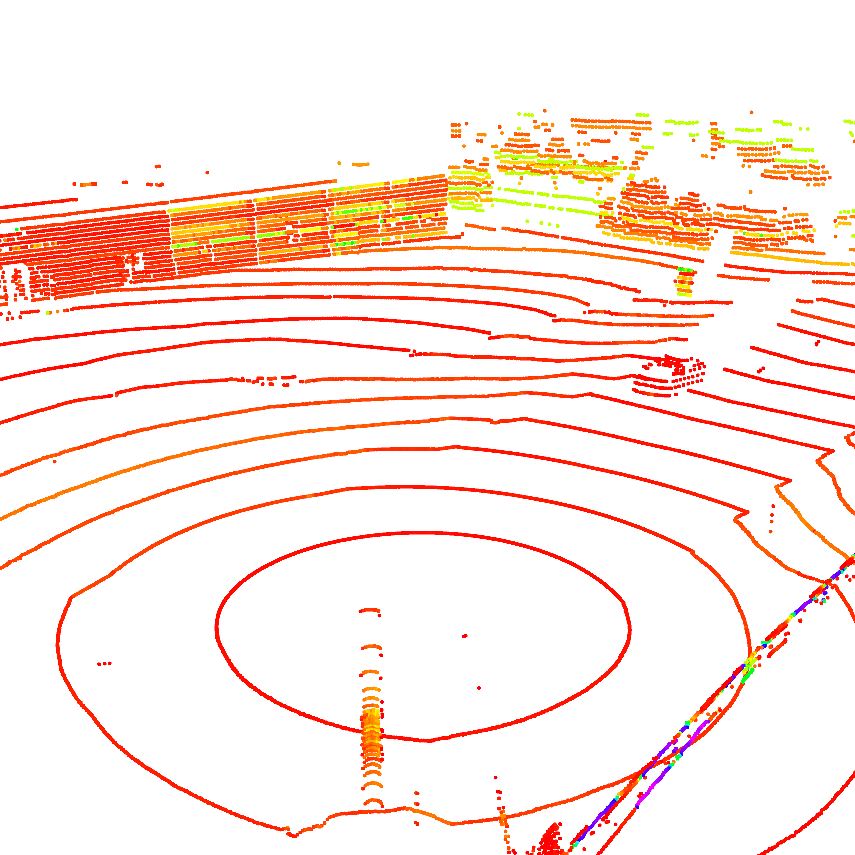}}
\label{fig:raw_data}
}
\subfigure[Feature-based: Road map.]{
\frame{\includegraphics[width=.225\textwidth]{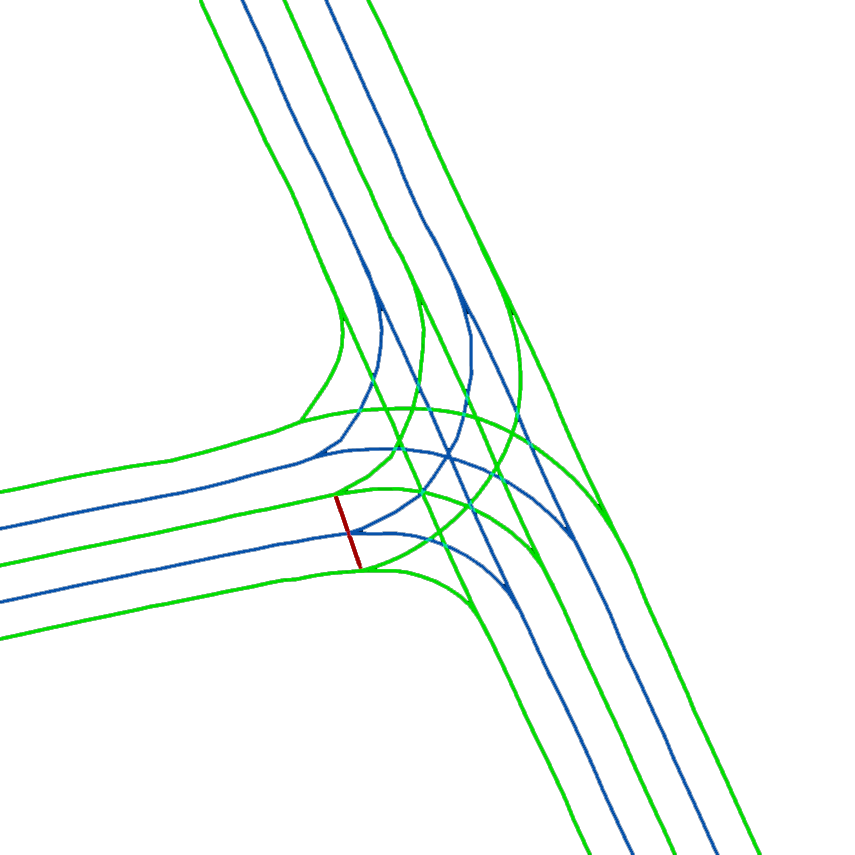}}
\label{fig:feature_based}
}
\subfigure[Grid-based: 2D occupancy grid.]{
\includegraphics[width=.225\textwidth]{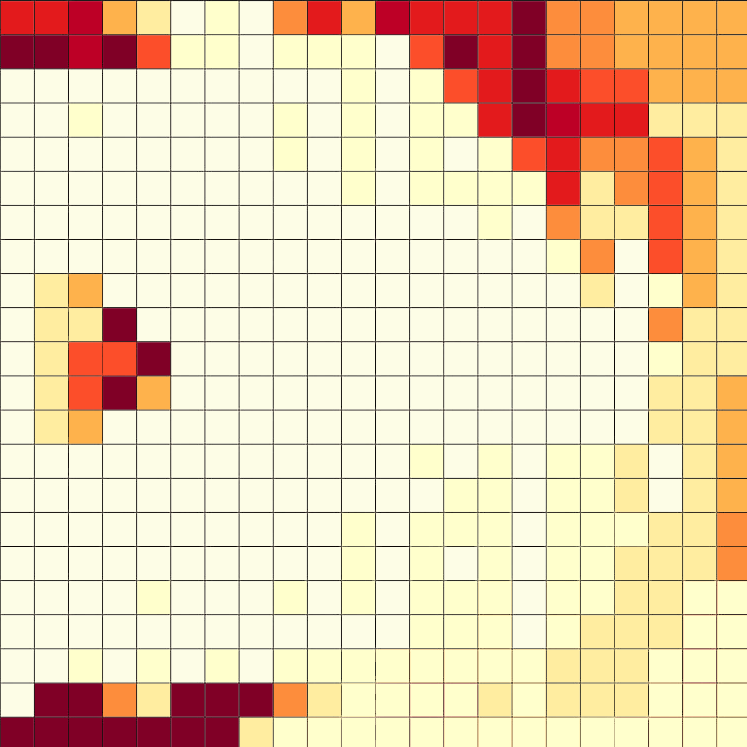}
\label{fig:grid_based}
}
\subfigure[Latent representation]{
\begin{minipage}{0.225\textwidth}
\vspace{-11em}
\hspace{2em}
\scriptsize
\begin{tikzpicture}[inner sep=0,minimum size=2.8mm,shape=circle]
\draw (0,0) node[draw,shape=rectangle](a) {\includegraphics[width=.5\textwidth]{figures/env_real.png}};
\draw (0,-1.7) node[shape=rectangle,minimum height=4.6mm](b){\textsf{encoder}};
\draw(-0.6,-1.45)--(0.6,-1.45) 
(-0.6,-1.45)--(-0.5,-1.95) 
(-0.5,-1.95)--(0.5,-1.95) 
(0.6,-1.45)--(0.5,-1.95); 
\draw (0,-2.6) node[draw,shape=rectangle,minimum width=0.6\textwidth,minimum height=4mm] (c) {};
\draw
(-1,-2.6) node[draw]{$v$}
(-0.5,-2.6) node[draw]{$w$}
(0,-2.6) node[draw]{$x$}
(0.5,-2.6) node[draw]{$y$}
(1,-2.6) node[draw]{$z$};
\draw[->](a)--(b);
\draw[->](b)--(c);
\end{tikzpicture}
\end{minipage}
\label{fig:latent_rep}
}

\caption{Four environment representations used in motion planning for autonomous driving, ordered from least to most abstract.}
\label{fig:env_rep}
\vspace{-5px}
\end{figure}

Human driver control actions are continuous, yet driving also contains discrete episodes, arising from road connectivity, signs, signals, road-user interactions, etc.
The vehicle must nevertheless follow a smooth continuous trajectory on the road.
The spectrum of possible discrete and continuous abstractions is thus the first principal axes of design space that we consider.
Some different representations derived from our autonomous driving stack are illustrated in Fig.~\ref{fig:env_rep}.

A common decomposition of the driving problem allocates high level discrete decisions ({\em go straight}, {\em turn left}, {\em stop}, \dots) to a behaviour planner (BP),
leaving low level continuous actions 
to be enacted by a local planner (LP).
Choosing discrete actions is well-suited to procedural programs, while continuous actions can be found by optimization.
In practice, the vehicle and external environment are not entirely predictable, such that high and low level control actions are dependent on each other, on the actions of other road users, and on the actual behaviour of the vehicle.
This leads to the second principal axis we consider, which concerns the overall architecture of the motion planner.
We consider the level of integration and communication between the behaviour planner and local planner, and the way prediction is incorporated, which we consider to have the most influence on BP design.

Our third and final principal design choice axis is the representation of decision logic.
In addition to conventional programs and expert systems, recent advances in hardware and software make learned logic a plausible and attractive alternative.

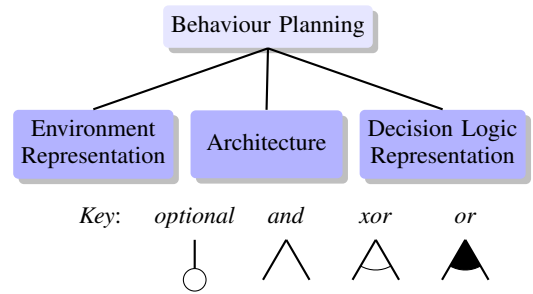
\begin{figure}[t]
    \vspace{-5em}
    \centering
    \small
    \begin{tikzpicture}
    \draw (0,0) node {
    \begin{forest}%
    for tree={parent anchor=south,
             child anchor=north,
             drop shadow,
             rounded corners=2pt,
             edge=thick,
             l+=0.7cm,
             fill=blue!30,
             delay={content={\strut #1}},
             align=center,
             base=center,
             minimum width=2cm
             },
    blackcircle/.style={tikz={\node[fill=black!60,inner sep=3pt,circle]at(.north){};}},
    whitecircle/.style={tikz={\node[draw,fill=white,inner sep=3pt,circle]at(.north){};}},
    [
    Behaviour Planning,fill=blue!10
      [Environment\\Representation]
      [Architecture,minimum height=0.9cm]
      [Decision Logic\\Representation]
    ]
    \end{forest}
     };
    \draw(0,0) node {
    \begin{forest}
    for tree={parent anchor=south,edge=thick,calign=fixed edge angles,calign angle=30,align=center},
    whitecircle/.style={tikz={\node[draw,fill=white,inner sep=3pt,circle]at(.north){};}},
    [
       [\textit{Key}:,no edge[,no edge]]
       [\textit{optional},no edge[,whitecircle]]
       [\textit{and},no edge[][]]
       [\textit{xor},no edge,name=O[,name=A][,name=B]]
       [\textit{or},no edge,name=V[,name=U][,name=W]]
    ]
    \draw (O.south) coordinate (Z) pic[draw,angle radius=4mm]{angle=A--Z--B};
    \draw (V.south) coordinate (Y) pic[draw,fill=black,angle radius=4mm]{angle=U--Y--W};
    \end{forest}
  };

  \end{tikzpicture}
    \vspace{-10pt}
    \caption{Feature diagram of high-level design choices. 
    }
    \label{fig:high-level}
    \label{fig:FDkey}
\vspace{-5px}
\end{figure}

Fig~\ref{fig:high-level} illustrates our three principal axes of design choices in the form of a feature diagram~\cite{KCHN90}.
The figure also contains a key to the visual syntax used elsewhere.

\smallskip

The remainder of the paper is structured as follows.
We first define the motion planning problem in Section~\ref{sec:problem}.
Then, in Section~\ref{sec:environment}, we consider the different ways that the motion planning environment can be represented, acknowledging our first axis of decomposition.
In Section~\ref{sec:architecture} we describe the different possible architectures of the BP, considering our second axis of decomposition. 
In Section~\ref{sec:decision} we outline the ways in which a BP's decision making can be implemented, which is our third axis of decomposition.
We conclude in Section~\ref{sec:conclusion}.

\section{Motion Planning Problem}\label{sec:problem}
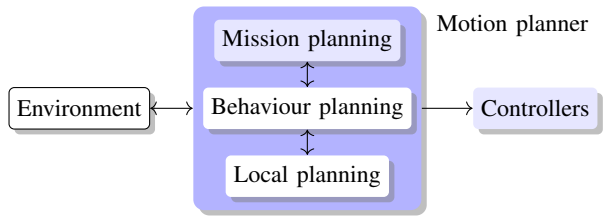
\begin{figure}[t]
\centering
\small
\begin{tikzpicture}[<->,box/.style={rectangle,minimum size=5.5mm,drop shadow,fill=white,rounded corners=2pt}]
\draw(0,0)node[draw,box](env){Environment};
\draw(3,0)node[box,fill=blue!30,rounded corners=4pt,minimum height=27mm,minimum width=30mm](mop){};
\draw[-](5.7,1.1)node{Motion planner};
\draw(3,0.9)node[box,fill=blue!10](mp){Mission planning};
\draw(3,0)node[box](bp){Behaviour planning};
\draw(3,-0.9)node[box](lp){Local planning};
\draw(6,0)node[box,fill=blue!10](cont){Controllers};
\draw(mp)--(bp);
\draw(bp)--(lp);
\draw(env)--(mop);
\draw[->](mop)--(cont);

\end{tikzpicture}

    \caption{General architecture of motion planning. White boxes denote the scope of the current work.}
    \label{fig:motionplanning}
    \vspace{-5px}
\end{figure}
The autonomous vehicle motion planning problem can be summarized as providing a reference trajectory, in continuous domain, that meets the specified requirements of safety, comfort, progress and energy efficiency, and which can be executed by the vehicle hardware, taking into account the vehicle's dynamics.
The task is typically decomposed into three sub-tasks: ({\em i}) mission planning; ({\em ii}) behaviour planning; and ({\em iii}) local planning.
These are illustrated in Fig.~\ref{fig:motionplanning}.
Mission planning is the process of making the highest level decisions with regard to the proposed journey, such as deciding the sequence of roads to take, given the desired destination, the current position of the vehicle, the user's preferences, and a priori assumptions about road conditions and availability.
If the mission planner's assumptions are contradicted along the way, or the user makes ad hoc changes to the requirements of the journey, the mission planner must re-plan.
Such decision making and re-planning are standard on current satellite navigation devices.
Since mission planning is already a well-discussed problem, we do not consider this aspect of motion planning in the present work.

Given a proposed sequence of roads to follow, as provided by the mission planner, the BP must devise a sequence of discrete high level control actions to navigate through the environment.
Control actions might include such basic manoeuvres as {\em speed up}, {\em slow down} and {\em come to a stop}. 
The actions must also be consistent with actual road conditions, so must be generated online according to sensor input.
Perception is, therefore, a crucial ingredient of behaviour planning with many of its own challenges, including noise, occlusions and sensor fusion.
In this work, we acknowledge the challenges of sensing by assuming that the motion planner is presented with a view of the ego vehicle and the external environment that may be incomplete and contain errors.
Notwithstanding the problems of imperfect assumptions and sensing, the BP must react to the dynamic nature of the environment. 

The objective of local planning is to devise a safe and smooth trajectory from the autonomous vehicle's current position to a goal position, by avoiding obstacles, satisfying comfort requirements and generally respecting kinodynamic constraints.
The abstract actions chosen by the BP must, therefore, take into account the actual state of the ego vehicle and environment.
The BP may parameterize its choices or otherwise communicate with the LP.
The optimal degree of integration between the LP and the BP depends on the algorithms chosen and how the overall task is divided, which will be discussd in Section~\ref{sec:architecture-mp}.


\section{Environment Representation}\label{sec:environment}

Design choices for the BP are often constrained by its input. This input is a representation of the driving environment modelled using information from on-board sensors and other sources, such as digital maps and Vehicle-To-Everything (V2X) communication.
The continuous spatio-temporal information about the environment is often discretized and abstracted before reaching the BP, to simplify it and pass on only the most relevant details to the processing pipeline~\cite{env_representation}. Such abstracted data requires reduced communication bandwidth and is more convenient to work with, aiding high frequency decision making.
On the other hand, the simplifying assumptions may be imperfect, and less-abstract data may contain useful additional information.
There thus exists a trade-off between the fidelity of decision-making and the computational burden.

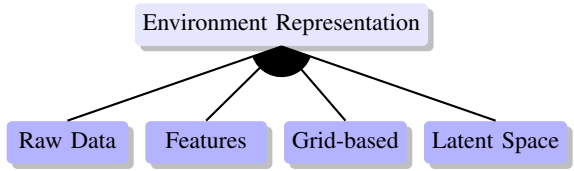
\begin{figure}
    \centering
    \small
    \begin{forest}%
    for tree={parent anchor=south,
             child anchor=north,
             drop shadow,
             rounded corners=2pt,
             edge=thick,
             l+=0.7cm,
             fill=blue!30,
             delay={content={\strut #1}},
             align=center,
             base=center,
             minimum width=1.6cm
             },
    blackcircle/.style={tikz={\node[fill=black!60,inner sep=3pt,circle]at(.north){};}},
    whitecircle/.style={tikz={\node[draw,fill=white,inner sep=3pt,circle]at(.north){};}},
    [Environment Representation,fill=blue!10,name=O
      [Raw Data,name=A]
      [Features]
      [Grid-based]
      [Latent Space,name=B]
    ]
    \coordinate (X) at (A.north);
    \coordinate (Y) at (B.north);
    \draw (O.south) coordinate (Z) pic[draw,fill=black,angle radius=4mm]{angle=X--Z--Y};
    \end{forest}
    \caption{Environment representation design choices.
    See Fig.~\ref{fig:FDkey} for key.}
    \label{figure:high_level_env}
    \vspace{-5px}
\end{figure}

The different types of representations are illustrated in the feature diagram in Fig.~\ref{figure:high_level_env} and discussed below, ordered from least to most abstract.
Although we treat them separately in the present context, we note here that a combination of these representations (as shown with the \emph{or} operator in the figure) may also be used in practice to provide a single coherent model of the environment \cite{Bansal2018Waymo}.

\subsection{Raw Data}
It is possible for the BP to use the direct output of on-board sensors and other sources (camera, radar, lidar, etc.), with minimal pre-processing.
Typical lidar output is illustrated in Fig.~\ref{fig:raw_data}.
Such output contains the most information and the most noise.
The complexity of making decisions with raw data makes this representation suitable only for end-to-end approaches, where the intermediate steps in the processing pipeline, such as road feature detection and path planning, are performed implicitly, e.g., by learning and encoding them into a neural network (NN). For instance, \cite{bojarski} transforms the raw pixels from a single camera sensor directly into steering commands using a convolutional neural network (CNN).
The resulting lack of modularity makes this approach less scalable and opaque to human understanding.

\subsection{Feature-based}
A set of features that encapsulate the current environment can be extracted from the raw sensor data. This set of features can be represented in discrete space~\cite{RB-NCS}, continuous space~\cite{Bender2014}, or a combination of both \cite{combiningneural}. This representation attempts to simplify the information in the scene, leading to a modular architecture design, while maintaining all the important information. However, errors within the feature representation can arise either due to bad feature selection or poor feature extraction, as this is a complex task.

In the autonomous driving domain, it is common to represent a geometric road network as a continuous feature (see Fig.~\ref{fig:feature_based}), which aids in the planning problem. An example of a discrete feature is a determination of whether a vehicle is at a stop sign or not. Finally, a bounding box is one of the most common features in autonomous driving, containing both continuous features, such as location and size, as well as discrete features, such as the object classification.

\subsection{Grid-based}
The feature-based representation can be further refined such that the continuous features are mapped to discrete space. The act of discretization significantly simplifies the information, allowing it to be conveniently used in traditional programmed logic systems. However, the resulting representation is less detailed, which can potentially lead to errors in decision making.

An occupancy grid is an example of such a grid-based environment representation. A 2D occupancy grid, as depicted in Fig.~\ref{fig:grid_based}, projects the road environment into the plane of the road surface and indicates the probability that a particular cell is occupied by an obstacle. The notion can be extended to 3D, e.g., voxel grids, and also equipped with object class and dynamics information to enable more subtle decision-making~\cite{env_representation}.

\subsection{Latent Representation}
Fig.~\ref{fig:latent_rep} illustrates how raw sensor data can be compressed into a latent variable representation of the environment.
While such a representation is typically more efficient, it may not be human-understandable.
On the other hand, the compression may capture patterns that are not apparent to a human programmer.
Typical encoding methods include Principal Component Analysis (PCA~\cite{Murphy2012}) and Variational AutoEncoders (VAE~\cite{ichter_harrison_pavone_2017}).

Since the input is not human-readable, the latent representation may be more amenable to data-driven planners. 
For example, a NN may create a condensed spatial and temporal representation of the environment that models not just the current state but also predicts future states, and a decoder can make this latent representation human-readable~\cite{world_models}.
The final latent representation captures all the relevant information in a small vector.
Some of these concepts are discussed further in Section~\ref{sec:learned}.

\section{Architecture}
\label{sec:architecture}

In this section, we discuss the possible ways in which a behaviour planner (BP) can be integrated with its associated modules, and how those choices influence the overall architecture of the planner.
The principal design choices are illustrated in Fig.~\ref{fig:arch_1}. The first design decision is to choose the right level of integration between BP and LP; the two planners can exist as separate or integrated into a single module. This choice is shown in the figure with an \emph{xor} operator under Motion Planner Architecture. The second design decision is whether to have a dedicated prediction module for BP (shown in the figure as an optional feature). One factor that influences this decision is the operational environment of the autonomous vehicle. For simple driving situations, where a constant velocity behaviour model for other road users might suffice, the architecture might not need a dedicated prediction module. However, for complex situations, especially in situations of high volume traffic, there might be a need for a dedicated prediction module to share the computation burden. In some designs, instead of prediction being a part of BP, the task can also be split between perception and LP. For example, the perception module can detect additional activity attributes (running, walking, stationary, etc.) as part of pedestrian detection, which LP can further consume to select a behaviour model to predict the trajectory of the road user. Another option is to not have prediction at all and make decisions using only the current world state, although this may not lead to correct decisions.

\subsection{Architecture of the Motion Planner}
\label{sec:architecture-mp}
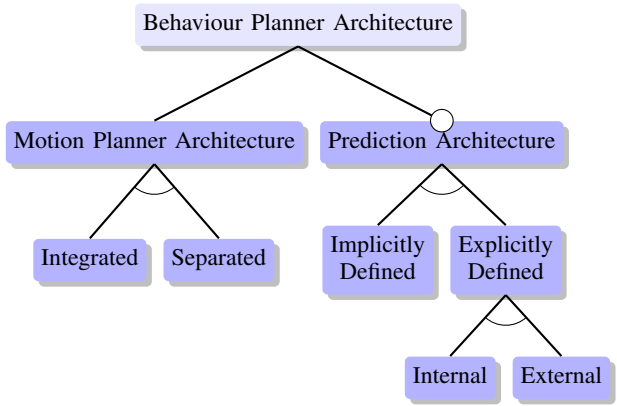
\begin{figure}
    \centering
   \small 
   \begin{forest}%
    for tree={parent anchor=south,
             child anchor=north,
             drop shadow,
             rounded corners=2pt,
             edge=thick,
             l+=0.7cm,
             fill=blue!30,
             delay={content={\strut #1}},
             base=center,
             align=center,
             },
    blackcircle/.style={tikz={\node[fill=black!60,inner sep=3pt,circle]at(.north){};}},
    whitecircle/.style={tikz={\node[draw,fill=white,inner sep=3pt,circle]at(.north){};}},
    [Behaviour Planner Architecture,fill=blue!10
      [Motion Planner Architecture
        [Integrated]
        [Separated]
      ]
      [Prediction Architecture, whitecircle
         [Implicitly\\Defined]
         [Explicitly\\Defined
            [Internal]
            [External]]
      ]
    ]
    \foreach \i/\j/\k in {!11/!1/!12,!21/!2/!22,!221/!22/!222}
    {
   \coordinate (A)at (\i.north);
    \coordinate (O)at (\j.south);
    \coordinate (B)at (\k.north);
    \path  (A)--(O)--(B)
    pic [draw, angle radius=4mm] {angle = A--O--B};
    }
    \end{forest}
\caption{Feature diagram of architecture design choices.
See Fig.~\ref{fig:FDkey} for key.}
    \label{fig:arch_1}
    \vspace{-5px}
\end{figure}

The integration of the BP within the motion planner is critical since this constrains the other elements and the overall architecture. As illustrated in Fig.~\ref{fig:arch_1}, there are two mutually exclusive methods of integration of the BP: one in which the BP module is architecturally separated from the rest of the motion planner, and one where the BP may be partially or fully integrated.\\

\subsubsection{Separated}

Many early approaches to motion planning \cite{ Urmson-2008-10020, bacha2008odin, urmson2007tartan } have a top-down architecture design, described in the problem definition and illustrated by the positions of motion planner elements in Fig.~\ref{fig:motionplanning}. In this architecture, the behaviour planning process is done on a planned mission path which is to be refined by a local planner (LP). The BP restricts the complexity of the local planning algorithm by providing a single or set of possible high-level manoeuvres at each step of the planning process. This decomposes the planning problem into two tasks, i.e. behaviour planning and local planning, with a clear separation between the two. Although there is an advantage of simplicity to this design, the choice of complete separation can lead to computational redundancy. For example, to avoid generating manoeuvres that cannot be safely executed by the LP, the BP might need to solve a part of the LP's trajectory planning problem. Even with the added computation, there can be conflicting solutions since the trajectory planning problem that the BP solves is typically less constrained than the one solved by the LP \cite{Urmson-2008-10020}.
 
Another approach inverts the traditional architecture for planning. This type of BP framework generates multiple possible local paths then selects the best manoeuvre according to a given cost function~\cite{wei2014behavioral}. This has the advantage of knowing the exact path that will be executed by the vehicle, and thus decisions are made on a more constrained problem, resulting in more accurate decisions. However, this approach results in increased computational redundancy as many paths now have to be considered.\\

\subsubsection{Integrated}
Recently, there has been a decline in human expert-based systems in favour of learned systems. Such learning based systems attempt to solve the entire motion planning problem \cite{learningdrivingstyles, combiningneural}, and perhaps even the entire autonomous driving problem, in an end-to-end fashion \cite{learninghowtodrive, bojarski, deepreinforcementlearningframework}. By encapsulating the motion planning problem in a single pipeline, manoevre selection can be performed implicitly during the planning process. To achieve this, the learned system must rely on rewards or cost functions to inform its behavioral decision-making; this can be done either with ~\cite{combiningneural} or without the use of explicitly defined manoevres ~\cite{Bansal2018Waymo}. While these approaches show great promise, they rely heavily on large labelled data sets and sophisticated simulation environments, both of which may not be readily available to academic researchers.

\subsection{Prediction Architecture}

In order to perform safe behaviour planning, many BP algorithms rely on predicting and reacting to the behaviour of other dynamic objects throughout the entire plan. Given their state information and past trajectories, the prediction task is to predict one or more of the following: the trajectory, the low-level motion primitives (accelerating, decelerating, and maintaining speed, etc.), or the intent (yielding, changing lane, crossing street, etc.).  

Environment prediction approaches vary according to the environment representation, the design of the prediction model, the abstraction of the prediction, the degree of incorporation of prior human knowledge, prediction horizon, robustness to noise, etc. \cite{gindele2015learning,zhan2016non,sarkar2017trajectory}. Despite this heterogeneity, three key categories of approaches stand out: ({\em i}) physics-based models, which predict dynamic object motion solely by physical laws; ({\em i}) manoeuvre-based models, which model the intended road-user manoeuvres and predict their execution; and ({\em i}) interaction-aware models, which account for the interdependencies among various agents in the environment \cite{lefevre2014survey}. Most published approaches are physics- or manoeuvre-based, with attention only recently shifting to interaction-based techniques.

The \emph{prediction architecture}, which is concerned with different levels of coupling between prediction and behaviour planning, is another key decision in the BP design space (Fig.~\ref{fig:arch_1}). The first choice is between the prediction models being \emph{explicitly} or \emph{implicitly} defined. \emph{Explicitly} defined prediction models take in state observations and make explicit predictions about the future behaviour of the road users. These models can be either \emph{external} or \emph{internal} to the BP.

\emph{External} prediction models are completely decoupled from the planning process, and their outputs augment the environment representation that is fed to the planner. This design choice offers a clear interface between prediction and planning, which aids modularity. Most prediction approaches fall into this category \cite{zhan2016non, gindele2015learning, sarkar2017trajectory, kafer2010recognition, koehler2013stationary, keller2014will, quintero2014pedestrian}.

\emph{Internal} prediction models, on the other hand, are integrated with the motion planning process and exist within the planner. Examples of this choice are planners that rely on a Partially-Observable Markov Decision Process (POMDP) model, which considers the intention of road users as a latent representation (see Section III-D) within the planning state space \cite{bandyopadhyay2013intention, wang2013probabilistic, bai2015intention, van2017motion}. At each planning step, the planner maintains a \emph{belief} of road users' intentions, which is periodically updated based on new observations. Planning over this belief space of road user intentions rather than using fixed predicted intentions from external models has advantages, especially with respect to planning under uncertainty. In particular, it may result in safer trajectories, especially in cases where road users change their intention in response to the subject vehicle's behaviour~\cite{bandyopadhyay2013intention}. However, this design is currently computationally intractable for situations with a high number of road users.

Instead of having explicitly defined prediction models, a BP architecture may use \emph{implicitly} defined ones, which results in an even higher degree of coupling between prediction and planning.  Implicit prediction models do not represent the intent of road users as explicit features. Instead, the prediction algorithm learns to predict the behaviour of road users simultaneously while learning a driving policy based on the interaction of the subject vehicle with the environment. Most approaches based on reinforcement learning (RL) fall under this category, which we elaborate further in Section~\ref{sec:learned}. The primary advantage of such a design is that it eliminates the need for a separate environment prediction module altogether, which reduces complexity and improves runtime efficiency. However, the human interpretability of implicit prediction models is lacking, which hinders their verification and validation.

\begin{figure} [t!]
    \centering
   \small
    \begin{forest}%
    for tree={parent anchor=south,
             child anchor=north,
             drop shadow,
             rounded corners=2pt,
             edge=thick,
             l+=0.7cm,
             fill=blue!30,
             align=center,
             base=center,
             delay={content={\strut #1}},
             },
    [Decision Logic Representation,fill=blue!10
      [Learned Logic
         [Learning from\\Example]
         [Learning from\\Interaction]
      ]
      [Programmed Logic
         [Declaratively\\Expressed
            [Expert Systems]
            [Optimization]]
         [Imperatively\\Expressed]
      ]
    ]
    \foreach \i/\j/\k in {!11/!1/!12}
    {
    \coordinate (A)at (\i.north);
    \coordinate (O)at (\j.south);
    \coordinate (B)at (\k.north);
    \path  (A)--(O)--(B)
    pic [angle radius=4mm] {angle = A--O--B};
    }
    \foreach \i/\j/\k in {!1/!0/!2}
    {
    \coordinate (A)at (\i.north);
    \coordinate (O)at (\j.south);
    \coordinate (B)at (\k.north);
    \path  (A)--(O)--(B)
    pic [draw, angle radius=4mm] {angle = A--O--B};
    }
    \end{forest}
    \caption{$\!$Feature diagram of decision logic design choices. See Fig.~\ref{fig:FDkey} for key.}
    \label{fig:arch}
    \vspace{-5px}
\end{figure}
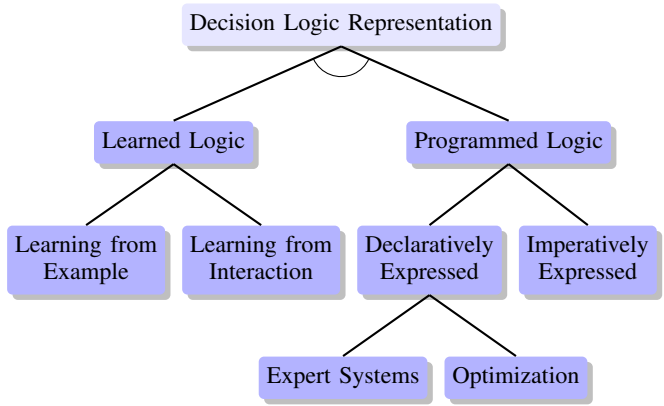

\section{Decision Logic Representation}
\label{sec:decision}

Our third principal axis of decomposition is that of the underlying logic representation used to make high-level decisions.
We categorize the decision logic of the planner into two paradigms: ({\em i}) logic represented through a set of explicitly programmed production rules, and ({\em ii}) logic representation relying on mathematical models with their parameters learned a priori. The final high level driving decision is made by exclusively following one of these two paradigms, as indicated by the highest level exclusive-or in Fig.~\ref{fig:arch}.

While \emph{learned-logic models} are able to generalize over diverse scenarios, they are not human-interpretable~\cite{lipton2016mythos} and hence, difficult for humans to ensure safety. On the other hand, \emph{programmed logic} requires extensive human effort and is subject to traditional software engineering principles. As a result, safety can be reasoned through traceability analysis.

\subsection{Programmed Logic}

Programmed logic systems have the advantage of interpretability which allows using traditional software engineering methods, such as inspections and walkthroughs to uncover problems in implementation. Unlike learned logic systems, they do not try to generalize over a set of examples and interactions but rather define consistent and verifiable resolution heuristics to arrive at a driving decision. The implementation of such systems can be achieved using two exclusively different programming paradigms as shown in Fig. \ref{fig:arch}: imperative and declarative.

Imperative systems outline a sequence of operations that expresses the control flow of a given program as it moves from one state to another. Declarative systems, on the other hand, are able to express the underlying logic without specifically describing the control flow. Due to their lack of strict system flow, declarative systems are more resistant to change since the system has fewer inter-dependencies, consequently making them harder to implement. Imperative systems, on the other hand, require a large number of rules and transitions in order to fully handle the complex cases of the driving task. As a result, this paradigm is hard to scale to more complex driving situations.\\

\subsubsection{Imperative Systems}

In an imperative system, the designers focus on sequences of operations that allow the transition from one state to another. Imperative functions work on concrete data and prepare the information flow required by the next function. The imperative programming paradigm brings about predictability as a result of well defined programmed instructions. Due to this reason, this paradigm is able to produce safety guarantees within a set of domain constraints.

State machines are an example of the imperative paradigm. Such systems exhaustively describe the list of all possible states as well as the transitions between them. One approach to design a state machine is to abstract the possible locations in the environment that can be encountered, and handle them independently~\cite{Urmson-2008-10020}. For each abstract location, the environment state causes the system to transition from one state to another, and finally produces a manoeuvre. For example, \cite{RB-LO} proposes six process flow diagrams allowing an autonomous vehicle to cross a multi-lane, controlled or uncontrolled intersection.
While being able to handle pre-defined driving scenarios, such systems are often less able to deal with noisy environment states.

A framework for composing multiple state machines is proposed in \cite{RB-MD}. This work reduces the exponential nature of the problem caused by a large number of transitions between states by using a rule-based system. It internally regulates the transitions to different state machines, each of which is responsible for selecting an appropriate manoeuvre.

In a more experimental context, the authors in \cite{RB-MMB} develop a state machine based solely on linear temporal logic (LTL) rules and show that a manoeuvre produced by such a system will not violate already existing rules. Their experiences show that this model of interaction between vehicles is sufficient for the ego to navigate without the risk of collision in a fully observable environment. However, this assumption of a fully observable environment is impossible to achieve in systems doing real world autonomous driving.

Although these systems show promise in the domain of safety, the authors do not believe that these approaches can adequately handle the noisy aspects of real world driving.\\

\subsubsection{Declarative Systems}

Declarative systems can represent the internal logic in the following two manners. ({\em i}) Expert systems obtain a decision through the evaluation of a world state over a set of rules and combine them through inferential processes to obtain a final behavioural decision. ({\em ii}) Optimization systems encapsulate driving behaviour as a set of mathematical variables which are optimized with respect to a notion of optimal behaviour.

Expert systems are able to derive a series of atomic propositions (AP) by combining them through an inferential process to arrive at a behaviour decision. For example, in \cite{RB-NCS}, the rules of the road are applied to a discretized world state represented by a set of boolean APs, which are then combined hierarchically. A hierarchical representation simplifies the rule premises and thus, improves generality and robustness of such a rule-base. Nevertheless, such severe discretizations yield only sub-optimal behavioural decisions as they rely on imprecise world representations.

To avoid the problems caused by discretization, a continuous world representation could be utilized. One approach to utilize this representation is through a probabilistic rule system. In \cite{RB-ACK}, the authors construct a probabilistic system and use a decision tree structure which operates on a continuous world state to produce a set of likelihood probabilities over all manoeuvres. Alternatively, a continuous world state can be processed using a set of fuzzy rules\cite{RB-JM}, which can handle noisy environmental states as well as a more complex set of scenarios.

Alternatively, the behavioural selection task can be solved by optimizing over a set of mathematically formulated criteria to achieve a correct behaviour. One approach to this type of optimization problem is the representation of the environment as an MDP. This allows the use of optimization solvers to select the path with the most optimal behaviour. Recent approaches use a POMDP representation model, which allows a world model to be constructed with incomplete world information~\cite{liu2015situation,wray2017online}. To make the solution tractable, the space of possible behaviours is usually discretized and decided in advance. The behaviours for the other agents can then be rolled out over time to predict potential future world states  ued to optimize the egos behaviour~\cite{Galceran2017}. MCTS can also be used to compute beliefs about the future world states~\cite{hubmann2018automated}. POMDPs have also been developed to work over continuous environment representations, which better models reality~\cite{Brechtel2014}. Recent advancements have enabled POMDP solvers to operate in real time, allowing deployment on autonomous vehicles~\cite{liu2015situation}.

\subsection{Learned Logic}\label{sec:learned}

Learned decision making can be categorized on the basis of whether the learning is from expert examples, or from interaction with a simulated environment that closely approximates the real world. There are also some approaches that use expert examples as well as a simulated interactive environment~\cite{ofjall2014biologically,visualautonomoussymbiotic,evolvinglarge,shalev2016safe}. In practice, ``learning from example'' has been shown to be more robust, although it requires extensive labelled driving data~\cite{Bansal2018Waymo}. On the other hand, ``learning from interaction'' allows for the acquisition of knowledge from a more diverse set of driving situations, even potentially dangerous situations.\\


\subsubsection{Learning from Example}\label{sec:from-example}
Simple \emph{end-to-end} representational learning has been shown to successfully execute basic manoeuvres on roads and highways~\cite{pomerleau1989alvinn, muller2006off, bojarski}. 
Since the driving task constitutes a large problem space, directly mapping sensor inputs to driving actions in this way needs a lot of data to learn effectively, meaning it is sample-inefficient. Due to the temporal nature of driving data, the authors~\cite{chen2018learning} demonstrated an improvement using deep recurrent NN over standard deep NNs. 


Another class of learning algorithms that use examples is \emph{learning from demonstrations}. A behaviour planning system can use \emph{imitation learning} to copy the behaviour of a reference driver~\cite{ziebart2008maximum,ross2011reduction, sun2018fast}. While behaviours learned through imitation learning are typically considered more robust in practice than other kinds of learned logic models, they often cannot adapt to scenarios which were sparsely represented in the training data. Conditional imitation learning~\cite{codevilla2018end} and adding noise to the samples~\cite{Bansal2018Waymo} can mitigate this. Recent work has established imitation learning as an interesting approach within this field.

It is also possible to use human demonstration examples to construct a reward function that can be used to learn a driving behaviour through RL~\cite{abbeel2008apprenticeship,silver2010learning} (see Section~\ref{sec:from-interaction} for RL). This process is called \emph{inverse} RL (IRL). The reward function must be postulated correctly (linear, non-linear, etc) before learning, otherwise the obtained reward function might not generalize to the driving preferences expressed by the examples. If designed properly, IRL algorithms result in driving behaviours that respect driving preferences specific to the human expert(s) who performed the demonstrations~\cite{learningdrivingstyles}. However, postulating a reward family has proven to be a difficult challenge and is still an open issue.\\

\subsubsection{Learning from Interaction}\label{sec:from-interaction}

A common way to learn from interaction is through RL, where an agent receives feedback from the simulated environment in the form of a reward and adjusts its behaviour to maximize the expected long term future reward~\cite{michels2005high,yu2017autonomous}. The two main ways to encapsulate the ego behaviour are using a set of discrete manoeuvres, or by using continuous control commands (steering angle, brake, and acceleration)~\cite{continuouscontrol,xiong2016combining}. Learning the continuous controls attempts to learn a solution to both the motion planning problem as well as learning the controls of the vehicle.

Learning driving behaviours with RL typically uses NNs, which suffer from ``catastrophic forgetting'', meaning they forget important previously learned knowledge when adding new knowledge. Mitigating catastrophic forgetting is an active area of research~\cite{kirkpatrick2017overcoming,lipton2016combating}.
The design of the reward function for RL is nontrivial, especially since there are usually multiple objectives, such as collision avoidance, comfort and traffic rule obedience. There are several ways to address this difficulty. The authors of~\cite{learninghowtodrive} incorporate a dependence on the immediate action in the reward design to improve the learned behaviour. Beyond immediate rewards, it is also possible to conveniently express complex driving objectives with temporal causality using temporal logic~\cite{littman2017environment,combiningneural,sadigh2014learning}.

The driving task can be also modelled as a hierarchical RL problem. In this formulation, the lowest level implementation of the hierarchy approximates a solution to the control and LP problem, while the higher level selects a maneuver to be executed by the lower level implementations~\cite{wei2014behavioral,safemultiagentrl,composingmetapolicies}. One advantage of such a decomposition is that the problem is divided into sub-problems (the lower for LP and the higher for BP) which are subsolvable and have lower individual variance~\cite{safemultiagentrl}. Empirical strategies like Monte Carlo Tree Search (MCTS) may be used to further improve the task of behaviour selection~\cite{combiningneural}. 


Another strategy is to use \emph{model-based} RL, in which specific characteristics about the environment model are learned. For example, the authors in \cite{uncertaintyaware} learn a collision probability model and in combination with RL methods, boost performance. With model-based RL, learning is more sample-efficient~\cite[Chapter~8]{sutton2018reinforcement}.

The state and action spaces can be made fuzzy~\cite{adaptivelearning}, which has the advantage of reducing the problem complexity and improving robustness. It is also possible to construct a driving behaviour which does not directly use the learned RL behaviour, but rather performs some probabilistic post inference~\cite{interactionawarehighways}. 
RL can also be done in a cooperative sense, by trying to maximize the system's utility rather than just a single (ego) vehicle's utility~\cite{individualvsdifference}; the system refers to the set of vehicles in the environment, all of which are following the learned behaviour.\\

\subsubsection{Learning from Example and Interaction}

Although ``learning from example'' and ``learning from interaction'' are the two broad categories within ``learned logic'' design choice, an approach may not be entirely one or the other section. For example, in~\cite{safemultiagentrl} an initial behaviour is obtained through imitation learning (learning from example), but improvements are made through interaction with a simulated environment (learning from interaction). While the hierarchical portion of this hybrid solution is still manual, the convergence in learning is faster since the initial behaviour is bootstrapped from an expert behaviour, and has to learn over a smaller problem space.

Biologically inspired methods such as Hebbian learning and genetic algorithms could also be combined with RL (learning from interaction) to learn a driving behaviour~\cite{ofjall2014biologically,visualautonomoussymbiotic,evolvinglarge}. For these methods to be effective, it is imperative to appropriately design the fitness function and spend the requisite time needed for parameter tuning. However, if done correctly, these methods may be able to produce multiple alternative optimal driving strategies. 

As discussed in Section \ref{sec:from-example} and  \ref{sec:from-interaction}, both ``learning from example" and ``learning from interaction" have advantages, like robustness and adherence to multiple objectives. If designed correctly then a combination can inherit the advantages of both approaches. However, if not appropriately designed, the resulting approach may be susceptible to the disadvantages highlighted in the previous sections. Nevertheless, the authors cautiously suggest that such an approach could be an interesting future direction of research.
\section{Conclusion}\label{sec:conclusion}
In this work we have systematized many feasible and practical design choices for behaviour planning in the context of autonomous driving.
There is an obvious trade-off between the representation of the world and the computational complexity of the motion planning problem, but
advances in hardware and software have made learned systems, especially in the field of neural networks, a feasible means to make decisions over continuous space.
Without learning, traditional robotic solutions cannot adequately handle complex, dynamic human environments, but ensuring the safety of learned systems remains a significant challenge.
Hence, although the present work has implied many crisp design choices, we speculate that future high performance and safe behaviour planning solutions will be hybrid and heterogeneous, incorporating modules consisting of learned systems supervised by programmed logic.

There is currently no benchmark to evaluate the performance of BP technologies, however a quantitative comparison remains important future work.

\bibliographystyle{IEEEtran}
\bibliography{IEEEabrv,bpdesignspace}

\begin{thebibliography}{10}
\providecommand{\url}[1]{#1}
\csname url@samestyle\endcsname
\providecommand{\newblock}{\relax}
\providecommand{\bibinfo}[2]{#2}
\providecommand{\BIBentrySTDinterwordspacing}{\spaceskip=0pt\relax}
\providecommand{\BIBentryALTinterwordstretchfactor}{4}
\providecommand{\BIBentryALTinterwordspacing}{\spaceskip=\fontdimen2\font plus
\BIBentryALTinterwordstretchfactor\fontdimen3\font minus
  \fontdimen4\font\relax}
\providecommand{\BIBforeignlanguage}[2]{{%
\expandafter\ifx\csname l@#1\endcsname\relax
\typeout{** WARNING: IEEEtran.bst: No hyphenation pattern has been}%
\typeout{** loaded for the language `#1'. Using the pattern for}%
\typeout{** the default language instead.}%
\else
\language=\csname l@#1\endcsname
\fi
#2}}
\providecommand{\BIBdecl}{\relax}
\BIBdecl

\bibitem{MYBM91}
A.~MacLean \emph{et~al.}, ``Questions, options, and criteria: Elements of
  design space analysis,'' \emph{Human–Computer Interaction}, vol.~6, no.
  3-4, pp. 201--250, 1991.

\bibitem{autonomoose100k}
\BIBentryALTinterwordspacing
T.~Pender, ``{W}aterloo's `{A}utonomoose' hits 100-kilometre milestone,''
  \emph{Waterloo Region Record}, August 2018. [Online]. Available:
  \url{www.therecord.com/news-story/8859691-waterloo-s-autonomoose-hits-100-kilometre-milestone/}
\BIBentrySTDinterwordspacing

\bibitem{KCHN90}
K.~C. Kang \emph{et~al.}, ``Feature-oriented domain analysis ({FODA}):
  feasibility study,'' Carnegie Mellon Univ., Software Eng. Inst., Tech. Rep.,
  1990.

\bibitem{env_representation}
M.~Schreier, ``Environment representations for automated on-road vehicles,''
  \emph{at - Automatisierungstechnik}, vol.~66, no.~2, pp. 107--118, feb 2018.

\bibitem{Bansal2018Waymo}
M.~Bansal \emph{et~al.}, ``Chauffeurnet: Learning to drive by imitating the
  best and synthesizing the worst,'' 2018, http://arxiv.org/abs/1812.03079.

\bibitem{bojarski}
M.~Bojarski \emph{et~al.}, ``End to end learning for self-driving cars,'' 2016,
  http://arxiv.org/abs/1604.07316.

\bibitem{RB-NCS}
N.~Zimmerman \emph{et~al.}, ``Implementing a rule-based system to represent
  decision criteria for on-road autonomous navigation,'' in \emph{2004 AAAI
  Spring Symp. on Knowledge Representation and Ontologies for Autonomous
  Systems}, 2004.

\bibitem{Bender2014}
P.~Bender \emph{et~al.}, ``Lanelets: Efficient map representation for
  autonomous driving,'' in \emph{2014 IEEE Intelligent Vehicles Symp. (IV)},
  June 2014, pp. 420--425.

\bibitem{combiningneural}
C.~Paxton \emph{et~al.}, ``Combining neural networks and tree search for task
  and motion planning in challenging environments,'' in \emph{2017 IEEE/RSJ
  Int. Conf. on Intelligent Robots and Systems (IROS)}, September 2017, pp.
  6059--6066.

\bibitem{Murphy2012}
K.~P. Murphy, \emph{Machine Learning, A Probabilistic Perspective}.\hskip 1em
  plus 0.5em minus 0.4em\relax MIT Press, 2012.

\bibitem{ichter_harrison_pavone_2017}
B.~Ichter \emph{et~al.}, ``Learning sampling distributions for robot motion
  planning,'' Sep 2017, http://arxiv.org/abs/1709.05448.

\bibitem{world_models}
D.~Ha and J.~Schmidhuber, ``World models,'' 2018,
  http://arxiv.org/abs/1803.10122.

\bibitem{Urmson-2008-10020}
C.~Urmson \emph{et~al.}, ``Autonomous driving in urban environments: {B}oss and
  the {U}rban {C}hallenge,'' \emph{Journal of Field Robotics Special Issue on
  the 2007 DARPA Urban Challenge, Part I}, vol.~25, no.~8, pp. 425--466, June
  2008.

\bibitem{bacha2008odin}
A.~Bacha \emph{et~al.}, ``{O}din: Team {V}ictor{T}ango's entry in the {DARPA}
  urban challenge,'' \emph{Journal of field Robotics}, vol.~25, no.~8, pp.
  467--492, 2008.

\bibitem{urmson2007tartan}
\BIBentryALTinterwordspacing
C.~Urmson \emph{et~al.}, ``{T}artan {R}acing: A multi-modal approach to the
  {DARPA} urban challenge,'' Carnegie Mellon Univ., Tech. Rep., 2007. [Online].
  Available: \url{http://repository.cmu.edu/robotics/967}
\BIBentrySTDinterwordspacing

\bibitem{wei2014behavioral}
J.~Wei \emph{et~al.}, ``A behavioral planning framework for autonomous
  driving,'' in \emph{IEEE Intelligent Vehicles Symp. Proc.}\hskip 1em plus
  0.5em minus 0.4em\relax IEEE, 2014, pp. 458--464.

\bibitem{learningdrivingstyles}
M.~Kuderer \emph{et~al.}, ``Learning driving styles for autonomous vehicles
  from demonstration,'' in \emph{2015 IEEE Int. Conf. on Robotics and
  Automation (ICRA)}, May 2015, pp. 2641--2646.

\bibitem{learninghowtodrive}
P.~Wolf \emph{et~al.}, ``Learning how to drive in a real world simulation with
  deep {Q}-networks,'' in \emph{2017 IEEE Intelligent Vehicles Symp. (IV)},
  June 2017, pp. 244--250.

\bibitem{deepreinforcementlearningframework}
A.~El~Sallab \emph{et~al.}, ``Deep reinforcement learning framework for
  autonomous driving,'' \emph{Electronic Imaging}, no.~19, pp. 70--76, 2017.

\bibitem{gindele2015learning}
T.~Gindele \emph{et~al.}, ``Learning driver behavior models from traffic
  observations for decision making and planning,'' \emph{IEEE Intelligent
  Transportation Systems Magazine}, vol.~7, no.~1, pp. 69--79, 2015.

\bibitem{zhan2016non}
W.~Zhan \emph{et~al.}, ``A non-conservatively defensive strategy for urban
  autonomous driving,'' in \emph{Intelligent Transportation Systems (ITSC),
  2016 IEEE 19th Int. Conf. on}.\hskip 1em plus 0.5em minus 0.4em\relax IEEE,
  2016, pp. 459--464.

\bibitem{sarkar2017trajectory}
A.~Sarkar \emph{et~al.}, ``Trajectory prediction of traffic agents at urban
  intersections through learned interactions,'' in \emph{2017 IEEE 20th Int.
  Conf. on Intelligent Transportation Systems (ITSC)}.\hskip 1em plus 0.5em
  minus 0.4em\relax IEEE, 2017, pp. 1--8.

\bibitem{lefevre2014survey}
S.~Lef{\`e}vre \emph{et~al.}, ``A survey on motion prediction and risk
  assessment for intelligent vehicles,'' \emph{ROBOMECH Journal}, vol.~1,
  no.~1, p.~1, Jul 2014.

\bibitem{kafer2010recognition}
E.~K{\"a}fer \emph{et~al.}, ``Recognition of situation classes at road
  intersections,'' in \emph{IEEE Int. Conf. on Robotics and Automation
  (ICRA)}.\hskip 1em plus 0.5em minus 0.4em\relax IEEE, 2010, pp. 3960--3965.

\bibitem{koehler2013stationary}
S.~Koehler \emph{et~al.}, ``Stationary detection of the pedestrian intention at
  intersections,'' \emph{IEEE Intelligent Transportation Systems Magazine},
  vol.~5, no.~4, pp. 87--99, 2013.

\bibitem{keller2014will}
C.~G. Keller and D.~M. Gavrila, ``Will the pedestrian cross? {A} study on
  pedestrian path prediction,'' \emph{IEEE Transactions on Intelligent
  Transportation Systems}, vol.~15, no.~2, pp. 494--506, 2014.

\bibitem{quintero2014pedestrian}
R.~Quintero \emph{et~al.}, ``Pedestrian path prediction using body language
  traits,'' in \emph{Intelligent Vehicles Symp. Proc., 2014 IEEE}.\hskip 1em
  plus 0.5em minus 0.4em\relax IEEE, 2014, pp. 317--323.

\bibitem{bandyopadhyay2013intention}
T.~Bandyopadhyay \emph{et~al.}, ``Intention-aware motion planning,'' in
  \emph{Algorithmic foundations of robotics X}.\hskip 1em plus 0.5em minus
  0.4em\relax Springer, 2013, pp. 475--491.

\bibitem{wang2013probabilistic}
Z.~Wang \emph{et~al.}, ``Probabilistic movement modeling for intention
  inference in human--robot interaction,'' \emph{Int. Journal of Robotics
  Research}, vol.~32, no.~7, pp. 841--858, 2013.

\bibitem{bai2015intention}
H.~Bai \emph{et~al.}, ``Intention-aware online {POMDP} planning for autonomous
  driving in a crowd,'' in \emph{2015 IEEE Int. Conf. on Robotics and
  Automation (ICRA)}.\hskip 1em plus 0.5em minus 0.4em\relax IEEE, 2015, pp.
  454--460.

\bibitem{van2017motion}
J.~Van Den~Berg \emph{et~al.}, ``Motion planning under uncertainty using
  differential dynamic programming in belief space,'' in \emph{Robotics
  Research}.\hskip 1em plus 0.5em minus 0.4em\relax Springer, 2017, pp.
  473--490.

\bibitem{lipton2016mythos}
Z.~C. Lipton, ``The mythos of model interpretability,'' 2016,
  http://arxiv.org/abs/1606.03490.

\bibitem{RB-LO}
Y.~Liu and U.~Ozguner, ``Human driver model and driver decision making for
  intersection driving,'' in \emph{2007 IEEE Intelligent Vehicles Symp. (IV)},
  June 2007, pp. 642--647.

\bibitem{RB-MD}
M.~Maurer and E.~D. Dickmanns, ``A system architecture for autonomous visual
  road vehicle guidance,'' in \emph{Proc. of Conf. on Intelligent
  Transportation Systems}, November 1997, pp. 578--583.

\bibitem{RB-MMB}
M.~Lahijanian \emph{et~al.}, ``Automatic deployment of autonomous cars in a
  robotic urban-like environment ({RULE}),'' in \emph{2009 IEEE Int. Conf. on
  Robotics and Automation (ICRA)}, May 2009, pp. 2055--2060.

\bibitem{RB-ACK}
M.~Ardelt \emph{et~al.}, ``Highly automated driving on freeways in real traffic
  using a probabilistic framework,'' \emph{IEEE Transactions on Intelligent
  Transportation Systems}, vol.~13, no.~4, pp. 1576--1585, December 2012.

\bibitem{RB-JM}
J.~P{\'e}rez~Rastelli and M.~Santos~Peñas, ``Fuzzy logic steering control of
  autonomous vehicles inside roundabouts,'' \emph{Applied Software Computer},
  vol.~35, no.~1, pp. 662--669, Jul 2015.

\bibitem{liu2015situation}
W.~Liu \emph{et~al.}, ``Situation-aware decision making for autonomous driving
  on urban road using online {POMDP},'' in \emph{Intelligent Vehicles Symp.
  (IV)}, 2015, pp. 1126--1133.

\bibitem{wray2017online}
K.~H. Wray \emph{et~al.}, ``Online decision-making for scalable autonomous
  systems,'' in \emph{26th International Joint Conference of Artificial
  Intelligence (IJCAI)}, 2017, pp. 4768--4774.

\bibitem{Galceran2017}
E.~Galceran \emph{et~al.}, ``Multipolicy decision-making for autonomous driving
  via changepoint-based behavior prediction: Theory and experiment,''
  \emph{Autonomous Robots}, vol.~41, no.~6, pp. 1367--1382, Aug 2017.

\bibitem{hubmann2018automated}
C.~Hubmann \emph{et~al.}, ``Automated driving in uncertain environments:
  Planning with interaction and uncertain maneuver prediction,'' \emph{IEEE
  Transactions on Intelligent Vehicles}, vol.~3, no.~1, pp. 5--17, 2018.

\bibitem{Brechtel2014}
S.~Brechtel \emph{et~al.}, ``Probabilistic decision-making under uncertainty
  for autonomous driving using continuous {POMDP}s,'' \emph{17th International
  IEEE Conference on Intelligent Transportation Systems (ITSC)}, pp. 392--399,
  Oct 2014.

\bibitem{ofjall2014biologically}
K.~{\"O}fj{\"a}ll and M.~Felsberg, ``Biologically inspired online learning of
  visual autonomous driving,'' in \emph{British Machine Vision Conf. 2014,
  Nottingham, UK September 1-5 2014}.\hskip 1em plus 0.5em minus 0.4em\relax
  BMVA Press, 2014, pp. 137--156.

\bibitem{visualautonomoussymbiotic}
K.~{\"O}fj{\"a}ll \emph{et~al.}, ``Visual autonomous road following by
  symbiotic online learning,'' in \emph{2016 IEEE Intelligent Vehicles Symp.
  (IV)}, June 2016, pp. 136--143.

\bibitem{evolvinglarge}
J.~Koutn\'{\i}k \emph{et~al.}, ``Evolving large-scale neural networks for
  vision-based reinforcement learning,'' in \emph{15th Annual Conf. on Genetic
  and Evolutionary Computation (GECCO)}.\hskip 1em plus 0.5em minus 0.4em\relax
  ACM, 2013, pp. 1061--1068.

\bibitem{shalev2016safe}
S.~Shalev-Shwartz \emph{et~al.}, ``Safe, multi-agent, reinforcement learning
  for autonomous driving,'' 2016, http://arxiv.org/abs/1610.03295.

\bibitem{pomerleau1989alvinn}
D.~A. Pomerleau, ``Alvinn: An autonomous land vehicle in a neural network,'' in
  \emph{Advances in neural information processing systems}, 1989, pp. 305--313.

\bibitem{muller2006off}
U.~Muller \emph{et~al.}, ``Off-road obstacle avoidance through end-to-end
  learning,'' in \emph{Advances in neural information processing systems},
  2006, pp. 739--746.

\bibitem{chen2018learning}
Y.~Chen \emph{et~al.}, ``Learning on-road visual control for self-driving
  vehicles with auxiliary tasks,'' 2018, http://arxiv.org/abs/1812.07760.

\bibitem{ziebart2008maximum}
B.~D. Ziebart \emph{et~al.}, ``Maximum entropy inverse reinforcement
  learning.'' in \emph{AAAI}, vol.~8.\hskip 1em plus 0.5em minus 0.4em\relax
  Chicago, IL, USA, 2008, pp. 1433--1438.

\bibitem{ross2011reduction}
S.~Ross \emph{et~al.}, ``A reduction of imitation learning and structured
  prediction to no-regret online learning,'' in \emph{14th Int. Conf. on
  artificial intelligence and statistics}, 2011, pp. 627--635.

\bibitem{sun2018fast}
L.~Sun \emph{et~al.}, ``A fast integrated planning and control framework for
  autonomous driving via imitation learning,'' in \emph{ASME 2018 Dynamic
  Systems and Control Conf.}\hskip 1em plus 0.5em minus 0.4em\relax American
  Society of Mechanical Engineers, 2018.

\bibitem{codevilla2018end}
F.~Codevilla \emph{et~al.}, ``End-to-end driving via conditional imitation
  learning,'' in \emph{2018 IEEE Int. Conf. on Robotics and Automation
  (ICRA)}.\hskip 1em plus 0.5em minus 0.4em\relax IEEE, 2018, pp. 1--9.

\bibitem{abbeel2008apprenticeship}
P.~Abbeel \emph{et~al.}, ``Apprenticeship learning for motion planning with
  application to parking lot navigation,'' in \emph{IEEE/RSJ Int. Conf. on
  Intelligent Robots and Systems (IROS)}.\hskip 1em plus 0.5em minus
  0.4em\relax IEEE, 2008, pp. 1083--1090.

\bibitem{silver2010learning}
D.~Silver \emph{et~al.}, ``Learning from demonstration for autonomous
  navigation in complex unstructured terrain,'' \emph{Int. Journal of Robotics
  Research}, vol.~29, no.~12, pp. 1565--1592, 2010.

\bibitem{michels2005high}
J.~Michels \emph{et~al.}, ``High speed obstacle avoidance using monocular
  vision and reinforcement learning,'' in \emph{22nd Int. Conf. on Machine
  Learning}.\hskip 1em plus 0.5em minus 0.4em\relax ACM, 2005, pp. 593--600.

\bibitem{yu2017autonomous}
L.~Yu \emph{et~al.}, ``Autonomous overtaking decision making of driverless bus
  based on deep {Q}-learning method,'' in \emph{2017 IEEE Int. Conf. on
  Robotics and Biomimetics (ROBIO)}.\hskip 1em plus 0.5em minus 0.4em\relax
  IEEE, 2017, pp. 2267--2272.

\bibitem{continuouscontrol}
T.~P. Lillicrap \emph{et~al.}, ``Continuous control with deep reinforcement
  learning,'' 2015, http://arxiv.org/abs/1509.02971.

\bibitem{xiong2016combining}
X.~Xiong \emph{et~al.}, ``Combining deep reinforcement learning and safety
  based control for autonomous driving,'' 2016,
  http://arxiv.org/abs/1612.00147.

\bibitem{kirkpatrick2017overcoming}
J.~Kirkpatrick \emph{et~al.}, ``Overcoming catastrophic forgetting in neural
  networks,'' \emph{National Academy of Sciences}, pp. 3521--3526, 2017.

\bibitem{lipton2016combating}
Z.~C. Lipton \emph{et~al.}, ``Combating reinforcement learning's {S}isyphean
  curse with intrinsic fear,'' 2016, http://arxiv.org/abs/1611.01211.

\bibitem{littman2017environment}
M.~L. Littman \emph{et~al.}, ``Environment-independent task specifications via
  {GLTL},'' 2017, http://arxiv.org/abs/1704.04341.

\bibitem{sadigh2014learning}
D.~Sadigh \emph{et~al.}, ``A learning based approach to control synthesis of
  {M}arkov decision processes for linear temporal logic specifications,'' in
  \emph{Decision and Control (CDC), 2014 IEEE 53rd Annual Conf. on}.\hskip 1em
  plus 0.5em minus 0.4em\relax IEEE, 2014, pp. 1091--1096.

\bibitem{safemultiagentrl}
S.~Shalev{-}Shwartz \emph{et~al.}, ``Safe, multi-agent, reinforcement learning
  for autonomous driving,'' 2016, http://arxiv.org/abs/1610.03295.

\bibitem{composingmetapolicies}
R.~Liaw \emph{et~al.}, ``Composing meta-policies for autonomous driving using
  hierarchical deep reinforcement learning,'' 2017,
  http://arxiv.org/abs/1711.01503.

\bibitem{uncertaintyaware}
G.~Kahn \emph{et~al.}, ``Uncertainty-aware reinforcement learning for collision
  avoidance,'' 2017, http://arxiv.org/abs/1702.01182.

\bibitem{sutton2018reinforcement}
R.~S. Sutton and A.~G. Barto, \emph{Reinforcement learning: An
  introduction}.\hskip 1em plus 0.5em minus 0.4em\relax MIT press, 2018.

\bibitem{adaptivelearning}
F.~Gritschneder \emph{et~al.}, ``Adaptive learning based on guided exploration
  for decision making at roundabouts,'' in \emph{2016 IEEE Intelligent Vehicles
  Symp. (IV)}, June 2016, pp. 433--440.

\bibitem{interactionawarehighways}
D.~S. Gonz{\'a}lez \emph{et~al.}, ``Interaction-aware driver maneuver inference
  in highways using realistic driver models,'' in \emph{2017 IEEE 20th Int.
  Conf. on Intelligent Transportation Systems (ITSC)}, October 2017, pp. 1--8.

\bibitem{individualvsdifference}
R.~Grunitzki \emph{et~al.}, ``Individual versus difference rewards on
  reinforcement learning for route choice,'' in \emph{2014 Brazilian Conf. on
  Intelligent Systems}, October 2014, pp. 253--258.

\end{thebibliography}

\end{document}